\documentclass[conference]{IEEEtran}
\IEEEoverridecommandlockouts
\usepackage{cite}
\usepackage{amsmath,amssymb,amsfonts}
\usepackage{graphicx}
\usepackage{textcomp}
\usepackage{xcolor}

\def\BibTeX{{\rm B\kern-.05em{\sc i\kern-.025em b}\kern-.08em
    T\kern-.1667em\lower.7ex\hbox{E}\kern-.125emX}}

\usepackage[locale = DE ]{siunitx}
\usepackage{acronym}

\usepackage{algorithm}
\usepackage[noend]{algpseudocode}
\usepackage{mathtools}

\newcommand\scalemath[2]{\scalebox{#1}{\mbox{\ensuremath{\displaystyle #2}}}}

\begin{document}

\title{Few-Shot System Identification for Reinforcement Learning
}

\author{\IEEEauthorblockN{Karim Farid}
\IEEEauthorblockA{\textit{Computer Science and Engineering Department} \\
\textit{American University in Cairo}\\
Cairo, Egypt \\
kfarid@aucegypt.edu}
\and
\IEEEauthorblockN{Nourhan Sakr}
\IEEEauthorblockA{\textit{Computer Science and Engineering Department} \\
\textit{American University in Cairo}\\
Cairo, Egypt \\
n.sakr@columbia.edu}

}

\maketitle

\begin{abstract}
Learning by interaction is the key to skill acquisition for most living organisms, which is formally called Reinforcement Learning (RL). RL is efficient in finding optimal policies for endowing complex systems with sophisticated behavior. All paradigms of RL utilize a system model for finding the optimal policy. Modeling dynamics can be done by formulating a mathematical model or system identification. Dynamic models are usually exposed to aleatoric and epistemic uncertainties that can divert the model from the one acquired and cause the RL algorithm to exhibit erroneous behavior. Accordingly, the RL process loses its generality because it is sensitive to operating conditions and changes in the model parameters. To address these problems, Intensive system identification for modeling purposes is needed for each system even if the model dynamics structure is the same, as the slight deviation in the model parameters can render the model useless in RL. The existence of an oracle that can adaptively predict the rest of the trajectory regardless of the uncertainties can help resolve the issue. The target of this work is to present a framework for facilitating the online system identification of different instances of the same dynamics class by learning a probability distribution of the dynamics conditioned on observed data with variational inference and show its reliability in robustly solving different instances of control problems without extra training in model-based RL with maximum sample efficiency.
\end{abstract}

\section{Introduction}
The availability of robust and scalable methods for system identification has multiple benefits in the framework of {\em reinforcement learning} \textsc{(RL)}, especially for systems that require further time and resources for each system individually. One of the benefits is the transformation facilitation of the learned policy from simulation to a real environment. The aforementioned procedure can be done by first modeling the dynamics within the real environment, then leveraging this dynamics model to either simulate data for planning (Dyna Q \cite{dyna}) or by directly planning through the model. This process lowers the overhead time and cost needed for system identification and its consequent modules. 

\par
\textsc{RL} research shows the reliability in learning strategies and policies for a vast range of planning and control tasks in different fields like finance \cite{RLECO}, games \cite{hideandseek, Dota, Rubik}, and robotics \cite{dqn} \cite{robo}. Many \textsc{RL} algorithms consist of three main steps: environment interaction, model learning, and behavior learning. First, the interaction with the environment and the generation of samples using a certain policy. Using a simulator or a  previously built model can accelerate this step to up to a thousand times faster than running on a real robot and adds the ability of a parallel generation of experience. The second step is the function inference from data. This step may fit a model to the transitions of the system to either the dynamics or estimate the returns of the Q-function \cite{dqn, robo}. Finally, the policy is improved via backpropagation through the model by using {\em model predictive control} (\textsc{MPC}) controllers \cite{diff-mpc} for finding an optimal policy,  or using the Q-function inferred to find the actions which maximize it. 

These steps are normally induced in iterations, where different techniques and repetitions at each step yield a different \textsc{RL} algorithm. Different types of \textsc{RL} algorithms with different trade-offs in sample efficiency, stability, and modeling assumptions for each algorithm emerge due to the flexibility in using different techniques in the steps of the framework. Sample efficiency demonstrates how fast an algorithm converges to a given reward, while stability measures the sensitivity of such convergence under different scenarios, i.e. using different hyperparameters or starting conditions.

\par
The \textsc{RL} framework may fit a pre-learned model into any of its three steps whether in model-free or model-based algorithms. The model directly affects model-based algorithms in which the cost over the time horizon is directly entangled with the model. However, in model-free algorithms, the model can be used to generate experience that resembles the actual system in a faster manner with no operative costs or safety hazards. A further discussion about the means for incorporating the model is presented in the third section.
\par
In this work, we aim to achieve the realization of few-shot system identification, which alleviates the burden of the identification process in \textsc{RL} frameworks while preserving and improving sample efficiency for models with the same dynamic structure. The few-shot nature of the model and the improvement in sample efficiency are attained by the oracle that finds the model with just the recent history of motion in an online manner. This is achieved by embedding the dynamic model by a neural network conditioned on limited-timesteps observations generated by the actual model through which the specific dynamic parameters can be digested. This model is required to have temporal consistency in contrast to single-step prediction networks using vanilla neural networks or gaussian processes\cite{GP} without memory or time overhead. In contrast to autoregressive models, in which both the dynamics and the states are encoded together in the hidden state \cite{ir}, our model learns them separately.

\par
We propose an approach that disentangles the dynamics from the states. We use a {\em recurrent neural network} (\textsc{RNN}) encoder to derive the dynamics as an {\em ordinary differential equation} (\textsc{ODE}) and a separate network for predicting the encoded states. By combining the current encoded states, the next control actions and the learned dynamics model in an ODE solver, we can predict the future states. 

We design two models: time-invariant system dynamics and the other for time-variant system dynamics. The main distinction between both models is their ability to change the dynamics efficiently with every new observation. After learning both models, they are employed in an optimal control problem using \textsc{MPC} to find optimal policies using the defined dynamics prediction model. 
The model is then evaluated intrinsically for predicting and extrapolating trajectories and extrinsically in a {\em model-based RL} (\textsc{MBRL}) system for finding an optimal sequence of actions with \textsc{MPC}, where optimization is carried in the latent space of the encoded state model. This approach can accommodate for learning optimal policies for different models with the same structure. In addition, this can be fit in almost all \textsc{RL} algorithms. Dynamic models can help accelerate the learning process concerning the wall clock time due to the ability of the trained model in simulation to scale well to real-world applications and close the reality gap by directly deploying models trained in simulation.

\section{Background and Related Work}\label{sec:Background and Related Work}

\subsection{Deep learning embeddings}
{\em Deep learning} (\textsc{DL}) uses embeddings to encode meaningful and complex features according to context in multiple fields of application. The work of \textsc{DL} embeddings is extended to the field of robotics in several ways. The most common way is finding the embeddings for high dimensional observations, such as for high dimensional raw images or videos for non-linear systems optimal control and model learning. In \cite{E2C}, the high-dimensional observation space is transformed to a lower dimension latent one that has local linear dynamics using variational inference rather than handing them to the model directly. Therefore, this form of transformation lowers the dimensionality of the observations to a concise state space for the model learning and the optimization problem. In this work, the probabilistic embeddings for the dynamics are predicted rather than for the observations to realize the goal of few-shot system identification by encoding the model structure and its basic parameters.

\subsection{Reinforcement learning}
\textsc{RL} is defined as learning to find the mapping of states to appropriate actions that achieve a certain behavior by optimizing a cost or reward function through interactions with the environment. In order to achieve this task, the system is usually formulated as a {\em Markov decision process} (\textsc{MDP}) with the goal of finding a policy $\pi$ to
control the Markov process under some objective. Given states $x \in \mathcal{X} $ and actions $u \in \mathcal{U}$
in the environment E, the (deterministic or stochastic) policy $u_{*} = \pi(x)$ maps states in $ \mathcal{X}$ to actions in $\mathcal{U}$ such that
$\pi^{\star}=\arg \max _{\pi} E_{\tau \sim p_{\pi}(\tau)}[ R_{t}]$, where  $R_{t}=\sum_{i=t}^{T} \gamma^{(i-t)} r\left(\boldsymbol{x}_{i}, \boldsymbol{u}_{i}\right)$ is
the return calculated according to a reward function $r(x,u)$. The objective is maximized under the distribution of a trajectory $\tau$ given by $p_{\pi}(\tau) = p_{\pi}\left(s_{1}, a_{1}, \ldots, s_{T}, a_{T}\right)$. The trajectory distribution is given by the initial state distribution $p(x_0)$ and a dynamics transition distribution $p(x_{t+1}|x_{t},u_{t})$ in the formula 
$p_{\pi}\left(x_{1}, u_{1}, \ldots, x_{T}, u_{T}\right)=\mathrm{p}\left(x_{1}\right) \prod_{t=1}^{T} \pi_{\pi}\left(u_{t} | x_{t}\right) p\left(x_{t+1} | x_{t}, u_{t}\right)$. The dynamics transition model can be deterministic or stochastic
\cite{rls}.

\subsection{Model-based reinforcement learning with a focus on system identification and transfer learning}

In this section, \textsc{MBRL} is reviewed as these methods are foundational to our work. There is a variety of algorithms used in \textsc{MBRL}, yet they all share their complete reliance on the transition model for cost optimization \cite{mbrls}. The counter model-free methods do not involve the model directly or rely on it for the cost calculation, but they implicitly include the model in the process of collecting experience.

\textsc{MBRL} is mainly a search method that is based on running simulations and evaluating the running rewards to learn the optimal policy. Although tree search methods are common in \textsc{MBRL} \cite{mcts, treesearch}, guided search methods are more suitable for stochastic optimization in order to find the optimal policy \cite{gps}.

The  {\em iterative linear-quadratic-Gaussian} \textsc{iLQG} \cite{ILQR-B} algorithm is an \textsc{MBRL} approach that solves the problem in a recursive and iterative manner by locally linearizing non-linear dynamics and approximating the cost to a quadratic function using Taylor series, and solving for the optimal policy using quadratic programming . The action-value function $Q(x_t,u_t)$ and the value function $V (x_t)$ are used in finding the gains for the linear feedback controller $g(x_t) = u_t + k_t + K_t(x_t - \hat{x_t})$, where $k_t$ is the open loop term, $K_t$ is the closed-loop feedback matrix, and $x_t$
and $u_t$ are the states and actions of the nominal trajectory.
Further improvements on the \textsc{iLQG} were attained through having more stability in learning, thus providing better convergence.
These improvements include adding line search for preventing overshooting, constraining changes in policy for continuously residing in trust regions, and guiding training with high entropy policies giving broad distribution with high rewards for avoiding local optima, and using guided samples for training as in \cite{trust-region, gps, MILQR}.

The previously mentioned algorithms can utilize fitted dynamics using different methods like Gaussian processes, neural networks, probabilistic neural networks, \textsc{RNN}s, ensembles, and Gaussian mixture models as in \cite{PDDM, IMIMFMB, pilco, E2C, gpsnn}. These fitted models can be incorporated through different means in \textsc{RL} algorithms. One approach is to imitate an optimal policy learned using these models as in \cite{IMIMFMB}. Another approach is to generate fictitious or imaginative roll-outs as in \cite{dyna, mba}, where the roll-outs serve as an off-policy experience, generated from other \textsc{MBRL} solvers, or an on-policy experience generated from the current policy on the learned model. The generalization of the model can be evaluated in an off-policy manner where the policy used in fitting the model is different from the one it is being evaluated on.

The work on \textsc{MBRL} also targets system identification, which is involved with dynamics model learning and transferring learned models across different domains, particularly from simulation to the real world \cite{real-sim-im, preparing}. Other efforts have been made in the learning process of \textsc{MBRL} for helping the convergence and shifting to the real world. For example, most \textsc{MBRL} frameworks would iteratively interleave the process for learning the system and learning the policy as in \cite{iterative}. Each iteration contributes data from a policy closer to the optimal one that benefits the modules of \textsc{MBRL} including the system identification module. This work is concerned with the \textsc{MBRL} process itself and the model learning approach. The nature of this work has been investigated in \cite{preparing}, in which an {\em online system identification} (\textsc{OSI}) process that predicts the model parameters on the fly feeds its prediction into a neural network {\em universal policy} (\textsc{UP}) that is parametrized by the predicted model parameters to find the optimal actions. Similar to the mentioned iterative process, the trajectory under the real dynamics aggregates the dataset for the \textsc{OSI} with the actions generated from the policy learned under the current \textsc{OSI}. However, In this work, there was no need for this iterative process which was essential in \cite{preparing}. Another addition in our \textsc{OSI} model is formulating the learning process in a more mathematically rigorous way by separately encoding the states and finding the distribution over the parameters using a generative model that linearly predicts the next encoded state.

\subsection{Variational Autoencoders}
The {\em variational autoencoders} (\textsc{VAE}s) provide \cite{VAE} a probabilistic spin on regular autoencoders \cite{AE}. Variational autoencoders are used to learn approximate posterior inference models for a variety of applications, e.g. recognition, de-noising, representation, compression, and visualization. The work in the robotics fields considers the usage of \textsc{VAE}s in finding semantically rich representation from high dimensional data as in \cite{rssm, E2C}. This approach can be changed from finding state embeddings to finding dynamic embeddings for achieving the purpose of this work. In the mentioned works, the \textsc{VAE} helps in finding the state posteriors inference models which are hard to evaluate analytically due to the non-linearity of the model dynamics and the high dimensionality of the states. The learning step is carried by optimizing the {\em evidence lower bound} \textsc{elbo} loss $\mathbf{E}_{z}\left|\log p_{\theta}\left(x^{(i)} | z\right)\right|-D_{K L}\left(q_{\phi}\left(z | x^{(i)}\right) \| p_{\theta}(z)\right)$. The \textsc{elbo} loss consists of the reconstruction part where the real data can be reconstructed from their latent space encodings and the KL-divergence part for making the approximate posterior distribution close to the prior. The variational inference idea is extended to sequential data with {\em variational recurrent neural networks} (\textsc{VRNN}s) \cite{VRNN} to grasp the stochasticity observed in highly structured sequential data such as natural speech. The latent space variables are not only stochastically dependent on the data shown but also the hidden states which remember the previous information. One of the significant differences in \textsc{VRNN} from the regular \textsc{VAE} is the use of priors conditioned on the previously seen steps to make use of the temporal structure. In \cite{WM}, an approach similar to \textsc{VRNN} is used to provide an imitation-learning-like algorithm that performs the behavior cloning task but in a probabilistic manner for the entire trajectory rather than the single steps. The ability to learn the probability distribution of the entire trajectory is essential in this work for time-variant and time-invariant systems and can be achieved in a similar manner as in \textsc{VRNN} with changing the latent space to represent the dynamics instead of the states.

\subsection{Neural Ordinary Differential Equations}
{\em Neural ordinary differential equations} (\textsc{NODE}) \cite{NODE}, the continuous counterpart of {\em residual networks} (\textsc{ResNets}), are used in finding continuous transitions over the network's depth space and used in multiple tasks, e.g. time series forecasting and image classification. 
\textsc{NODE} parameterizes the differential equation controlling the flow of the states in the form of a neural network with parameters $\theta$  where $\frac{\mathrm{d} \mathbf{z}(t)}{\mathrm{d} t}=\mathbf{f_{\theta}}(\mathbf{z}(t), t)$.
The parameters $\mathbf{\theta}$ are optimized using an objective $\mathcal{L}_{\theta}(z)$. 
The formula carries a strong resemblance to dynamic systems in the real world as most robotic and industrial systems are modeled in this form. Therefore, the approach for learning the \textsc{ODE} should be suitable in system identification in which time-variant non-linear models can be learned easily. The work in \cite{NODE} also includes a generative modeling approach for time series which is extended in this work to controlled trajectories that condition the predictions on the future control inputs. All relevant information in the trajectory, including the dynamics and the states, is packed in a single vector $\mathbf{z}$ rather than explicitly decoupling the dynamics and states.

\section{Method for Online System Identification}\label{sec:Method}
In this section, the design of the oracle that predicts the future states is developed and is later incorporated in an \textsc{MBRL} algorithm  for solving classic control tasks. To address the aforementioned problems, system identification is carried for modeling purposes but needed to be done for each system. Given the oracle, few-shot system identification can be realized using domain randomized data \cite{DomRand}. The target of the next two sub-sections is elaborate on the method for finding the parameters $\mathbf{A_t, B_t, o_t, \sigma_t}$ which govern the following distribution \begin{equation}
\mathbf{x_{t}}\ \sim N(\mathbf{A_t} \mathbf{x_{t-1}}+\mathbf{B_t} \mathbf{u_{t-1}} + \mathbf{o_t}, \mathbf{\sigma_t})
\end{equation}

Where,  at any given time $t$, $\mathbf{A_t}$ is the linear system matrix for the encoded state $\mathbf{x_t}$, $\mathbf{B_t}$ is the input matrix that takes into account the effect of the control action, $\mathbf{o_t}$ is an offset, and $\mathbf{\sigma_t}$ is the standard deviation of the above distribution.

This equation is solved for the parameters in time-invariant dynamics, where the parameters are constant with time, and time-variant dynamics. To facilitate the learning of the mentioned, the model uses an encoded feature $z_{t}$ for feature $X_{t}$. The nature of the model enforces the conversion of the dynamics from nonlinear to linear in the encoded feature space. The methodology of learning the dynamics or transformation matrix is used in multiple domains as in \cite{PN} for neutralizing the effect of affine transformations, which does not have an effect on 3D object detection and segmentation. The dynamics uses with an ODE solver to predict the future points in the trajectory conditioned on the control inputs and the initial state. With a robust predictive model, \textsc{MBRL} can adapt, without relearning the model, to provide a generic optimal policy learning method. The implicit linearization feature in the model benefits the process of finding the optimal control solution as most of the algorithms need a linearization step to find the optimal policy. 

\subsection{VCNODETI: Variational controlled neural ODE for controlled time-invariant models}

In this section, we develop the approach for learning time-invariant dynamics embeddings with the help of an \textsc{RNN} encoder for the mean and variance of the dynamics parameters. The \textsc{ODE} solver uses the predicted dynamics, the initial state and control actions to make predictions into the future.

The model finds the embedding of the actual dynamics of the system given a few trajectory data points. These embeddings are used thereafter as the predictive model. A novel architecture that formulates the predictive model towards generic \textsc{MBRL} is proposed, which is inspired by \cite{NODE}. The model is split into three main parts: the state encoder, the dynamics model, and the state decoder. The states are encoded using a neural network $\phi^{enc}$  as a flexible feature extractor. The importance of finding relevant feature encodings is to act as a kernel that lifts the features from a space where the dynamics is non-linear into a space of linear dynamics. The second part is a generic \textsc{RNN} encoder for a general class of dynamics which given the previously observed points within a specified window yields an embedding vector that encodes the prevalent dynamics within that trajectory. The embedding dynamics vector is then restructured into a $dxd$ transformation matrix $A$ and $dxu$ control input matrix $B$, where the $d$ is the dimension of the new feature space of $\mathbf{Z}$ and  $u$ is the dimension of the action space. In order to stabilize the learning process, the network predicts the deviation around the identity matrix rather than directly predicting the transformation matrix itself. A possible approach towards dimensionality reduction for the transformation matrix prediction from $dxd$ to $2d$ is the replacement of the mentioned prediction output with a rank-one-matrix as the deviation from the identity matrix through predicting two d-dimensional vectors $v$ and $r$ where $ \boldsymbol{T= I + rv^T}$ as in \cite{E2C}. The final part is the decoder network $\phi^{dec}$ which transforms the features from the encoded space back to their original one. Optionally, the dynamics generated from the reparameterization trick from the vector inferred by the \textsc{RNN} dynamics encoder is passed through a network $\phi^{dynamics}$ as an intermediate step. In order to account for the disturbance and non-linearity of the data, the predictive model is trained in a similar manner as with the \textsc{VRNN} using the \textsc{ELBO}-loss. The sequence of the dynamics prediction model is as follows

\begin{equation}
\begin{aligned}
\mathbf{z}_{t_{1}}, \mathbf{z}_{t_{2}}, \ldots, \mathbf{z}_{t_{N}} &= \phi^{enc}\left(\mathbf{x}_{t_{1}}, \mathbf{x}_{t_{2}}, \ldots, \mathbf{x}_{t_{N}}\right) \\
\mathbf{d_u, d_{\sigma}} &=\operatorname{RNNEncoder}\left((\mathbf{z,u})_{t_1 \rightarrow_N} \right) \\
\mathbf{d} & \sim \mathcal{N}\left(\mathbf{d_u, d_{\sigma}}\right)\\
\mathbf{d} & = \phi^{dynamics}\left(\mathbf{d}\right)\\
\mathbf{z}_{t_{N+1}}, \mathbf{z}_{t_{N+2}}, \ldots, \mathbf{z}_{t_{M}}  &=\operatorname{ODESolve}\left(\mathbf{z}_{t_{N}}, d, \mathbf{T,U}\right) \\
\text { each } \mathbf{x}_{t_{i}} & \sim \phi^{dec}\left(\mathbf{x} \mid \mathbf{z}_{t_{i}}\right)
\end{aligned}
\end{equation}
and then minimizing the \textsc{ELBO} loss,

\begin{equation}
\begin{aligned}
\sum_{i=N+1}^{M} \log p(\mathbf{x}_{t_{i}} \mid \mathbf{x}_{t_{i-1}}, \mathbf{d})+\log p(\mathbf{d})-\log q(\mathbf{d} \mid {\mathbf{X_{1 \rightarrow N}}})
\end{aligned}
\end{equation}
In the equation above,  $p(\mathbf{d})$ follows a uniform distribution, where $p(\mathbf{d}) \sim \mathcal{N}(0,\mathbf{I})$.
The model leverages the ability to solve \textsc{ODE}s in both directions of time and at continuous timesteps. In contrast to the approach in \cite{NODE}, the model separates the prediction of the dynamics from encoding the states.  The model achieves $0.5-3 \% $ {\em relative root mean squared error} (\textsc{RMSE}) for non-linear dynamics models. The inference time of the recurrent dynamics encoder and the state encoder and decoder is almost negligible compared to the time complexity of the Dormand-Prince \textsc{ODE}-solver (ODE45) \cite{ODE45}.

\subsection{VCNODET: Variational controlled neural \textsc{ODE} for controlled time-variant models}
\begin{figure}[h]
\centering
\includegraphics[scale = 2, width=0.48\textwidth]{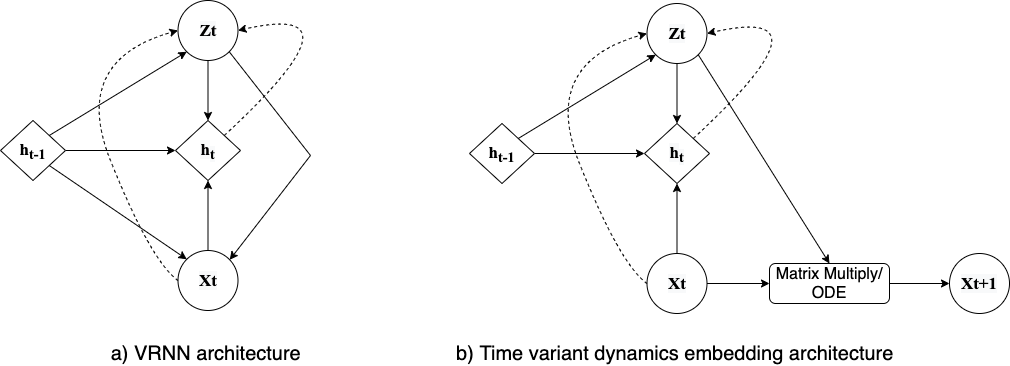}
\caption{Time variant dynamics embedding Architecture}
\label{fig:VCNODET}
\end{figure}
To extend the model to the time-variant dynamics, it needs to predict the dynamics in every timestep. This approach resembles the \textsc{VRNN} method applied in speech recognition and handwritten text generation \cite{VRNN}. The target of the \textsc{VRNN} is to encode the states in the more representative latent space and also to embed the temporal relation exhibited in the data. After training the \textsc{VRNN} model, the generation is carried by sampling the latent space variables inferred from the prior network conditioned on the previous hidden state (in the case of the initial step it would be zero hidden states). Then, the latent space features are used along with the hidden states to generate the original state and then finding the next hidden state and so forth.

The difference in this model is the reliance on transformation matrices to get the next state from the current one and using the \textsc{VRNN} to encode the changing dynamics. This approach handles the case of time-variant dynamics due to its capability of changing the encoded dynamics online. The previous hidden state is used to compute the probability of the latent dynamics variable using the prior network. The inference network for the dynamics is conditioned on the current hidden state and the input. Similar to the time-invariant model, the next state is found by using a transformation matrix, which represents the \textsc{ODE} governing the transitions, and the encoded previous state. The model is trained in a similar manner to \textsc{VRNN} with maximizing the likelihood of the data, as the changes are done implicitly in the prediction model and the architecture. The corresponding loss to the maximum likelihood is 

\begin{equation}
\scalemath{0.85}{
\sum_{t=1}^{T} [
\mathbf{KL}(q(z_{t} \mid x_{\leq t}, z_{<t}) \mid p(z_{t} \mid x_{<t}, z_{<t})) \\ -\log p(x_{t} \mid z_{\leq t}, x_{<t})]}
\end{equation}

\subsection{Model Predictive Control with VCNODE}

The model developed is utilized in a {\em quadratic programming} (\textsc{QP}) context to solve for the optimal policy which can be solved in a \textsc{MPC} paradigm. The following optimization problem in the equation below formulates the task, which is to develop a policy that can reach the target state optimally. The target of the optimal control problem is the trade-off between constraints and costs. In the case of using the states directly, rather than the encoded ones, the dynamics would be non-linear and would incur an extra burden for finding the optimal policy. Therefore, the lift of the states to the encoded feature space with linear dynamics is suitable for the task. The cost equation $\mathbf{C}$ is used as a measure of the deviation from the optimal states and the magnitude of the actions taken. To calculate the cost and constraints, the target states and boundaries are also lifted to the encoded feature space of the proposed models. This change of variables in the control problem avoids the repeated linearization and added steps for the cost, constraints, and boundaries. Hence, the formulation of the control problem would be

\begin{equation}
\scalemath{0.85}
{\begin{array}{lcl}
\min & \left[c_{T}\left(\mathbf{z}_{T}, \mathbf{u}_{T}\right)+\sum_{t_{0}}^{T-1} c\left(\mathbf{z}_{t}, \mathbf{u}_{t}\right)\right] & \\
\text { s.t. }& z_{0}=z & \\
& z_{k+1}=\left(A_{k} z_{k} + B_{k} u_{k} + o_t\right), & k=0, \ldots, N-1 \\

& \underline{z} \leq z_{k} \leq \bar{z}, & k=1, \ldots, N \\
& \underline{u} \leq u_{k} \leq \bar{u}, & k=0, \ldots, N-1
\end{array}
}
\end{equation}

In order to link between the accuracy of the predictive models and the optimal control algorithm, the training of the network is optimized to estimate up to N timesteps which corresponds to the time horizon of the \textsc{MPC} controller solver.

\section{Experimental Results}\label{sec:Experimental Results}

\subsection{Predictions on a toy dataset}

The models are evaluated on fitting and extrapolating the trajectories of different random systems with the same structure. The state-space model of the data is given by $\mathbf{\mathcal{\dot{X}}_{t}} = \begin{pmatrix}
  W_{11} & W_{12}\\ 
  W_{21} & W_{22}
\end{pmatrix} \mathbf{\mathcal{X}_{t}}$. Each of the parameters of the \textbf{W} Matrix is sampled from the random normal distribution $\mathbf{\mathcal{N}}(0,0.5)$. W11 and W22 are scaled by -0.1 and shifted by -0.05. With equal probabilities, W21 or W12 are scaled by -1.0. This scaling and shifting procedure yields a spiral form in the feature space to match the one in \cite{NODE}. A total of 12000 trajectories, split on an 80/20 basis for the train/test datasets, are generated flowing from time 0 to 35 with a total of 400 points in between. The trajectories are then windowed with a size of 61. The first half is used for predicting the dynamics and the second half is then inferred given the predicted dynamics as in the figure below. To facilitate the training, random timesteps (one for each trajectory) within the first half of the window are chosen as the starting points for the inferred trajectories, and teacher forcing is used with a decreasing probability with the training epochs. All neural networks used, including the state encoder, state decoder, and the network inferring the actual dynamic parameters from the latent dynamics generated by the \textsc{RNN} dynamics encoder, have 4 hidden layers with 128 units and {\em rectified linear units} (\textsc{ReLUs}) as activation functions. The dynamics encoder used is a 2 layered bidirectional \textsc{LSTM} \cite{LSTM} with states of dimension 32. The dynamics is solved with an Euler solver for real-time performance, however, to attain better accuracy, more accurate solvers can be used. The data is augmented by adding Gaussian noise to the states.

\begin{figure}[h]
\centering

\includegraphics[width=0.495\textwidth, height= 0.455\textwidth]{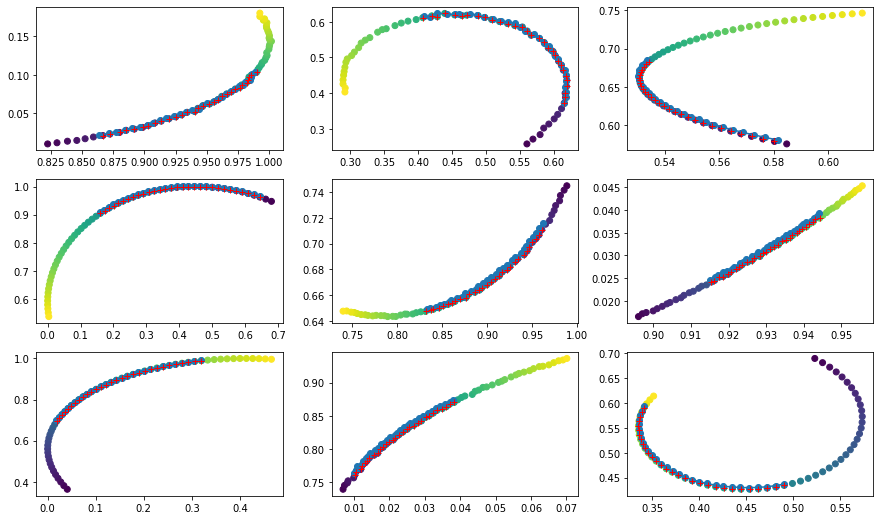}
\caption{Toy dataset prediction. The red pluses represent the points to be predicted in the main trajectory. The blue plot represents the predicted trajectory. the first half of the window starting from the dark to the light yellow color is used to infer the dynamics. }
\label{fig:Toy dataset prediction}
\end{figure}

\subsection{OpenAI classic control dataset}

The model is evaluated extrinsically and intrinsically on the pendulum swing-up and cart-pole environment in OpenAI gym for predicting the trajectory and using it as a predictive model in \textsc{MBRL}. \cite{gym}.

100,000 episodes of 500 timesteps for the given systems with dynamics randomization are generated. The parameters are sampled with a standard deviation of 30$\%$ from the mean values independently out of a normal distribution. In the case of the pendulum, mean values are the gravity of 10 $m/sec^2$, the pole length of $1m$, and the mass of 1 $kg$. These values of the parameters are from the original pendulum swing-up problem. Costs are incurred on the squared amount of $\theta$ which is the pole angle in addition to the regularization for the magnitude of the angular speed and control action according to $.1 * {\omega}^2 + .001 * u^2$.  The speed is clamped at 8 $rad/sec$ in both directions. The prediction of the dynamics is carried on a window of 61 timesteps. The first 30 is dedicated to the inference of the dynamics and the last 30 are the ones to be predicted. The control actions for the transition data creation are generated with random values within the applicable control ranges, in which two methods are used in data generation: completely random values and Perlin noise \cite{pnoise}. Perlin noise was more suitable for exploring the state space where a sequence of pseudo-random numbers was generated at randomly chosen scales. The use of Perlin noise as a random control action signal provides a random temporal structure for better exploration of the state space rather than the regular random number generator that lacks structure in time and performs weakly in terms of exploration.

\subsubsection{Architecture and Training details}
All features are normalized with min-max normalization. The architecture is similar to the one in the toy example except for the addition of the inference of the control input matrix, $\mathbf{B}$, used for the control action. No encoding is carried on the control action to facilitate the optimal control process. 

\subsubsection{Results}
The results for the proposed approach are evaluated for real-time performance, convergence to the optimal states, and the total cost incurred for the pendulum task which is nonlinear in nature. Speed clamping is added as another means of non-linearity. 10,000 different instances are sampled according to the aforementioned distribution and are solved with the \textsc{MPC} optimal controller with a time horizon of 24 steps. To decouple the \textsc{MPC} performance from the model, all 10,000 instances are successfully planned with the true dynamics with an average running cost of -211. Regarding the real-time performance, the time required for the system without a solver is on average 7 ms, while the main overhead is due to the ODE solver. In the case of using ODE45 solver, it takes 60ms on average while using Euler's solver takes 3 ms.

\begin{figure}[ht!]
        \begin{center}
               
         \includegraphics[width=0.49\textwidth, height= 0.55\textwidth]{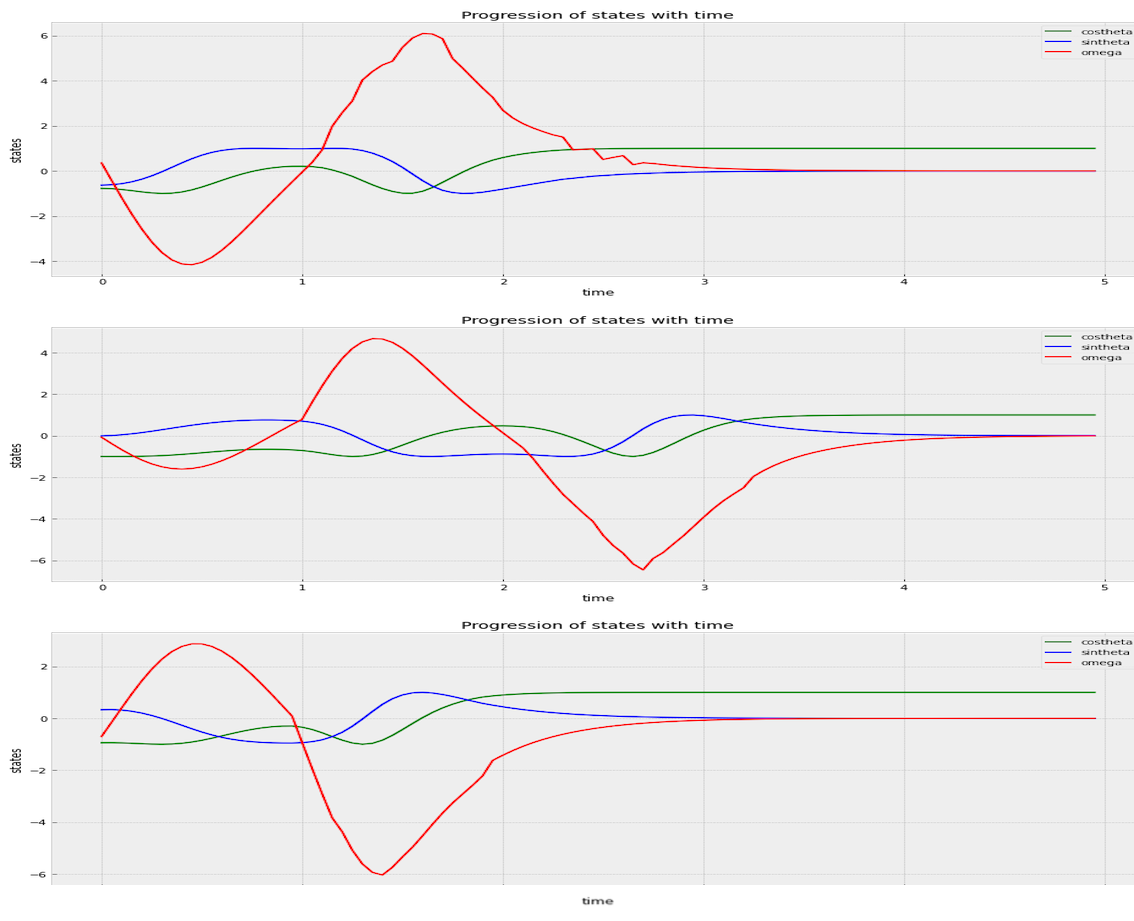}
         \caption{progression of states with time for the pendulum task, note the convergence to the optimal states. Omega is expressed in $rad/sec$.}
        \label{fig:pendulum states}
        \end{center}  
    \end{figure}
    
\begin{table}[ht!]
\centering
\caption{Pendulum swing up (Underscore means divergence).}
\begin{tabular}{|c|c|c|c|}
\hline
           & \textbf{RMSE}   & \textbf{Average cost} & \textbf{Success rate} \\ \hline
CNODE-only & 0.00754         & \_                    & \_                    \\ \hline
CNODE-VAE  & 0.142           & \_                    & \_                    \\ \hline
Ours       & \textbf{0.0016} & \textbf{-231}         & \textbf{95.32\%}      \\ \hline
\end{tabular}

\label{table:cost evaluation}
\end{table}


\section{Conclusion and Future Work}
This work proposes a reliable framework for the realization of few-shot system identification using conditional generative models for learning \textsc{ODE}s as an oracle that can generically predict the behavior for various systems within the same structure through making use of the temporal relation in these systems. The experiments target the intrinsic and extrinsic evaluation of the model. The intrinsic evaluation of the \textsc{RMSE} for recreating the trajectory and extrapolating the predictions from the inferred linear transformation matrix and the encoded previous states proves the reliability of the proposed approach. The model is extrinsically evaluated through solving optimal control problems on nonlinear systems within \textsc{QP} context with \textsc{MPC} where the trajectory optimization is carried out on the latent space of the dynamics embeddings model. The results show the robustness of such a framework in system identification for reinforcement learning and that the target of lowering the burden for system identification is attained. The work is planned to be evaluated on more robust \textsc{ODE} solvers such as ODE45.
Future work includes:
\begin{itemize}

\item Scaling to the real world, high dimensional applications, and the assessment of the reality gap divergence.  
\item Using \textsc{RNN-ODE} for embedding the dynamics \cite{ir} and embedding high dimensional sequential observations like videos using a separate temporal encoder for the states.

\item Enforcing closer embedding for trajectories from the same system by imposing a lifted structure loss. \cite{skilltransfer}

\end{itemize}



\end{document}